\documentclass[conference]{IEEEtran}
\IEEEoverridecommandlockouts
\usepackage{cite}
\usepackage{amsmath,amssymb,amsfonts}
\usepackage{algorithm}
\usepackage{algorithmic}
\usepackage{graphicx}
\usepackage{textcomp}
\usepackage{xcolor}
\usepackage{multirow}
\usepackage{hyperref}
\def\BibTeX{{\rm B\kern-.05em{\sc i\kern-.025em b}\kern-.08em
    T\kern-.1667em\lower.7ex\hbox{E}\kern-.125emX}}
\DeclareRobustCommand*{\IEEEauthorrefmark}[1]{%
    \raisebox{0pt}[0pt][0pt]{\textsuperscript{\footnotesize\ensuremath{#1}}}}

\begin{document}

\title{Latent Feature and Attention Dual Erasure Attack against Multi-View Diffusion Models \\for 3D Assets Protection}

\author{
\IEEEauthorblockN{
Jingwei Sun\IEEEauthorrefmark{1},
Xuchong Zhang\IEEEauthorrefmark{1,*},\thanks{*Correspondence to Xuchong Zhang (zhangxc0329@xjtu.edu.cn)}
Changfeng Sun\IEEEauthorrefmark{1},
Qicheng Bai\IEEEauthorrefmark{1},
Hongbin Sun\IEEEauthorrefmark{1}}
\IEEEauthorblockA{\IEEEauthorrefmark{1}National Key Laboratory of Human-Machine Hybrid Augmented Intelligence, \\National Engineering Research Center of Visual Information and Applications, \\Institute of Artificial Intelligence and Robotics, Xi'an Jiaotong University, Xi’an, 710049, China}

\IEEEauthorblockA{\{310412, sunchangfeng, navelOrange\}@stu.xjtu.edu.cn, zhangxc0329@xjtu.edu.cn, hsun@mail.xjtu.edu.cn}}

\maketitle

\begin{abstract}
Multi-View Diffusion Models (MVDMs) enable remarkable improvements in the field of 3D geometric reconstruction, but the issue regarding intellectual property has received increasing attention due to unauthorized imitation. Recently, some works have utilized adversarial attacks to protect copyright. However, all these works focus on single-image generation tasks which only need to consider the inner feature of images. Previous methods are inefficient in attacking MVDMs because they lack the consideration of disrupting the geometric and visual consistency among the generated multi-view images. This paper is the first to address the intellectual property infringement issue arising from MVDMs. Accordingly, we propose a novel latent feature and attention dual erasure attack to disrupt the distribution of latent feature and the consistency across the generated images from multi-view and multi-domain simultaneously. The experiments conducted on SOTA MVDMs indicate that our approach achieves superior performances in terms of attack effectiveness, transferability, and robustness against defense methods. Therefore, this paper provides an efficient solution to protect 3D assets from MVDMs-based 3D geometry reconstruction. The code is publicly available at: \href{https://github.com/super-jw/LFADEA}{\textcolor{magenta}{https://github.com/super-jw/LFADEA}}
\end{abstract}

\begin{IEEEkeywords}
Multi-View Diffusion Models, adversarial attacks, 3D assets protection
\end{IEEEkeywords}

\section{Introduction}
\label{sec:intro}
3D geometric reconstruction is an important task in computer graphics~\cite{charatan2024pixelsplat, wu2024reconfusion}, providing fundamental support for applications such as video games and virtual reality. Recently, with the introduction of diffusion models~\cite{croitoru2023diffusion, yang2023diffusion}, remarkable advancements have emerged in this field and multi-view images can be precisely generated to reconstruct 3D geometric in a few seconds.
Although this brings great conveniences, the issue regarding intellectual property has received increasing attention, because malicious users can imitate the 3D structures of any product by leveraging its sample images from the internet even without permission, thereby gaining illegal profits. Therefore, it is necessary to develop effective intellectual property protection technologies to prevent the theft of 3D assets.
\begin{figure}[!t]
\centering
\includegraphics[width=0.8\linewidth]{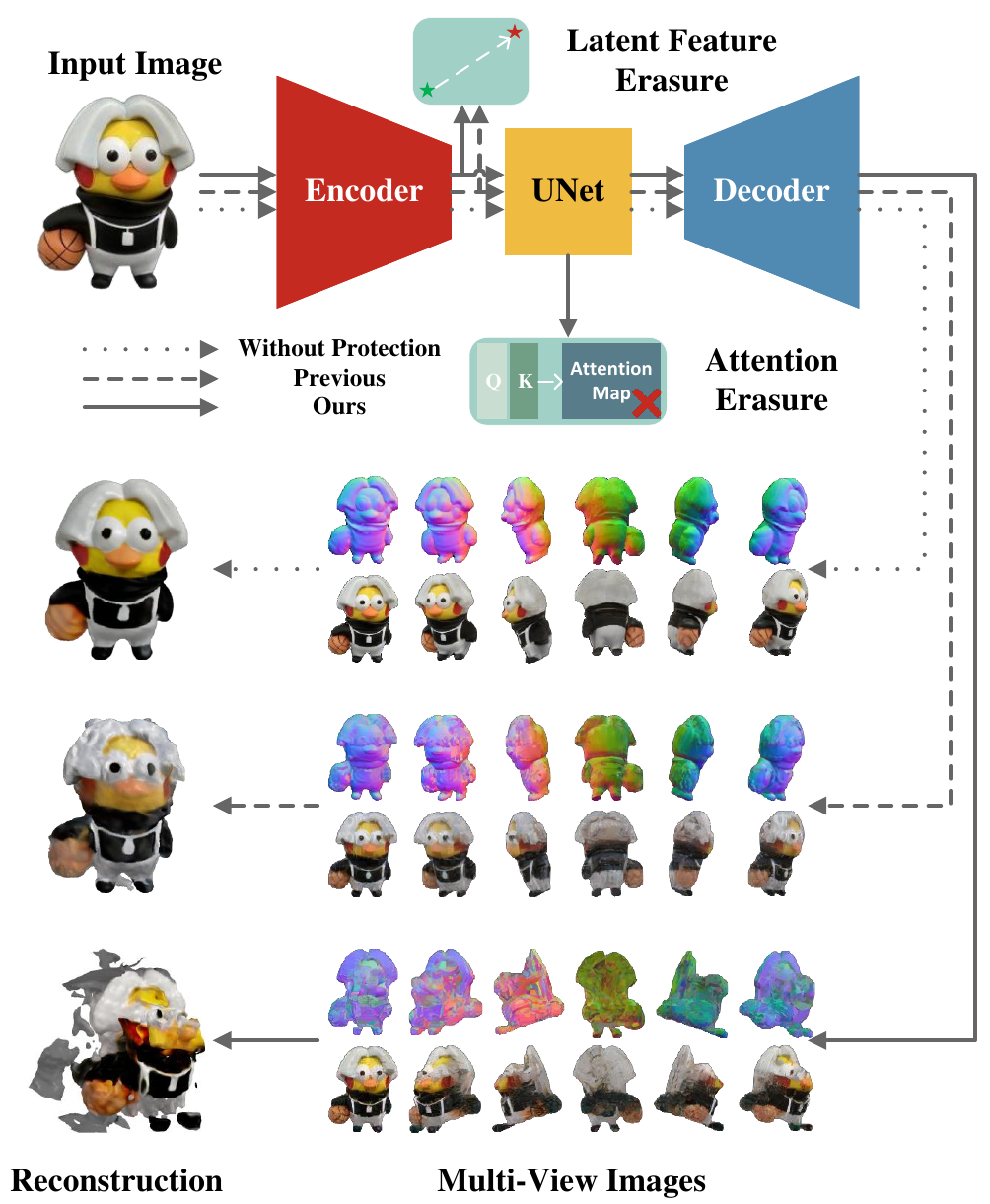}
\caption{
Comparison of reconstruction results between our method and previous methods. The first row: without protection, 3D geometry can be reconstructed accurately. The second row: a previous method WAE~\cite{zhu2024watermark} perturbed latent feature, resulting in chaotic content, but the reconstructed outline remains complete. The third row: our method erasures both latent feature and attention, significantly degrading 3D reconstruction quality.
}
\label{figure1}
\end{figure}

\begin{figure*}[t!]
\centering
\includegraphics[width=0.9\linewidth]{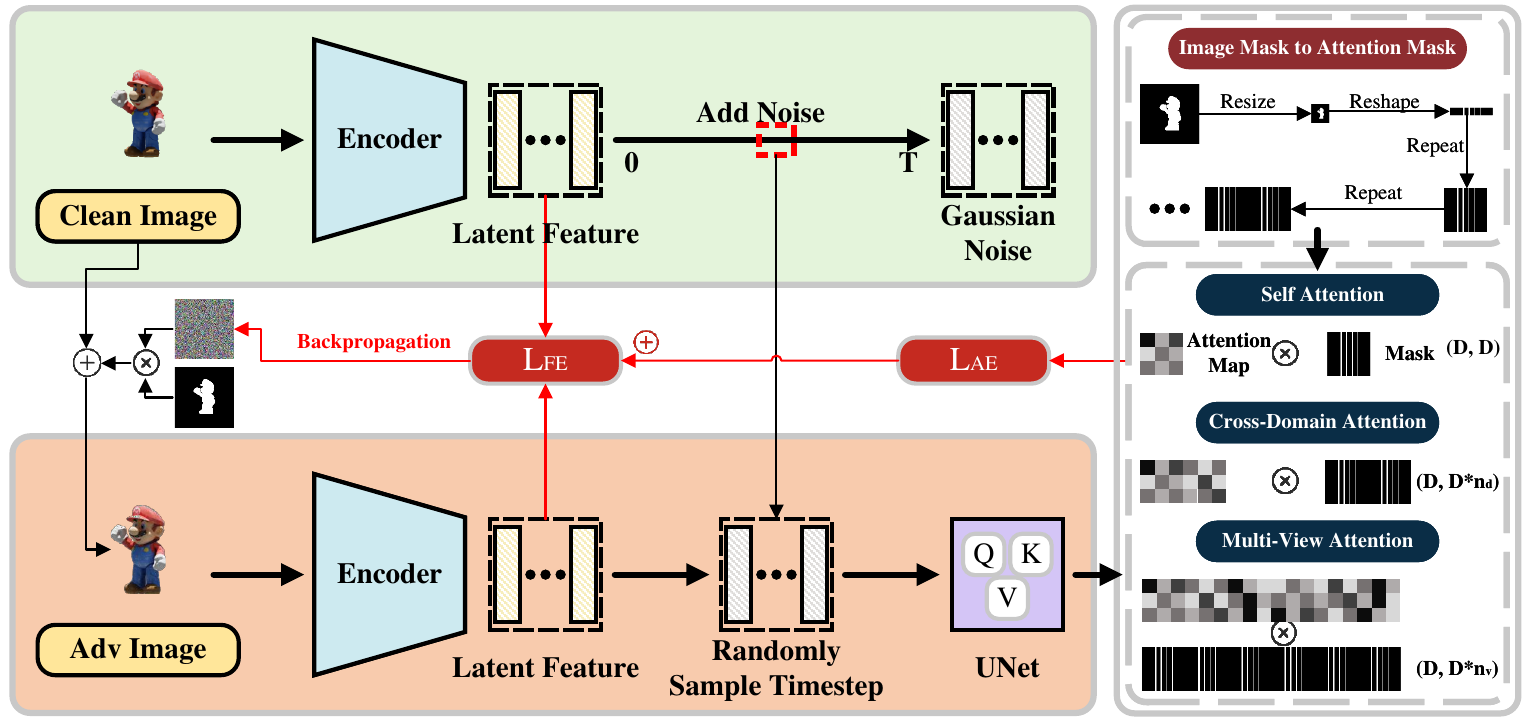}
\caption{
The overall framework of the proposed method. Firstly, we extract the latent feature of the adversarial image and design a latent feature erasure loss $L_{FE}$ to drive it away from the distribution of the clean image. Secondly, we randomly sample timestep in each iteration and establish an attention erasure loss $L_{AE}$ to divert the attention of the region of interest to other regions, thereby disrupting the geometric and visual consistency among the generated multi-view
images. Eventually, we combine $L_{FE}$ and $L_{AE}$ to form the final dual erasure loss, then the perturbation can be updated by the gradient descent algorithm.
}
\label{figure2}
\end{figure*}

Multi-View Diffusion Models (MVDMs) are a type of generative model widely used in 3D geometric reconstruction. The early works~\cite{liu2023zero, chan2023generative} utilize stable diffusion to generate multi-view images from a single-view image. However, these methods mainly use self-attention which considers only a single view at each iteration, resulting in the generation of 3D shapes
 exhibiting inconsistencies~\cite{long2024wonder3d}. Subsequently, multi-view attention~\cite{shi2023mvdream, liu2023syncdreamer} and cross-domain attention~\cite{long2024wonder3d, pang2024envision3d} are introduced to stable diffusion framework to improve the consistency and quality of generated images. According to our extensive investigation, adversarial attacks against diffusion models show great potential to prevent unauthorized image generation. All of the current attack methods~\cite{liang2023adversarial, liang2023mist, zhu2024watermark} focus on single image generation tasks which only need to consider the inner feature of images, such as DreamBooth~\cite{ruiz2023dreambooth}, LoRA~\cite{hu2021lora}, and Custom Diffusion~\cite{kumari2023multi}. Therefore, they optimize adversarial examples to disrupt the latent features' distribution of the input image in order to degrade the quality of the generated images. However, the attack on latent features is insufficient in multi-view images generation task, because it lacks the consideration of disrupting the geometric and visual consistency across the generated multi-view images. As shown in the second row of Figure~\ref{figure1}, directly transferring previous attack method (WAE) to MVDMs leads to some chaotic effects by disrupting the distribution of latent features, but the outline is still complete since the consistency between multi-view images has not been disturbed.

Apart from perturbing the distribution of latent features, we claim attacking the attention mechanisms in MVDMs can disrupt the geometric and visual consistency of the generated multi-view images effectively. In this paper, we propose a latent feature and attention dual erasure attack method against MVDMs. Specifically,
we take a full consideration of various attention mechanisms in MVDMs, including self-attention, multi-view attention, and cross-domain attention mechanisms, and then propose an identical attention erasure loss to reduce the attention of regions of interest. In this manner, the attention of MVDMs will be diverted from regions of interest (i.e., foreground regions) to background regions, thereby disrupting the consistency among the generated multi-view images. Moreover, we also combine an additional feature erasure loss to make the latent features deviate from the original distribution to achieve the proposed dual erasure attack. The experiments conducted on two SOTA MVDMs and Google Scanned
Objects dataset indicate that our approach achieves superior performances in terms of attack effectiveness, transferability, and robustness against defense methods. Our main contributions are summarized as follows:
\begin{itemize}
\item[$\bullet$] To our best knowledge, we are the first to consider intellectual property infringement issues arising from MVDMs and use adversarial attacks to protect 3D assets.
\item[$\bullet$] We propose a novel latent feature and attention dual erasure attack to disrupt the inner feature and consistency across the generated multi-view images simultaneously.
\item[$\bullet$] The extensive experiments demonstrate that the proposed method can greatly degrade the 3D reconstruction quality from multi-view images, compared to previous methods.
\end{itemize}

\section{Method}

% In this section, we first introduce the preliminaries about MVDMS. Then we describe the proposed latent feature and attention dual erasure attack method in detail.
\subsection{Preliminaries and Overall Framework}
\label{preliminaries}
Unlike 2D diffusion models that generate a single image, MVDMs aim to simultaneously generate multiple images from different views, denoted as $p_m{(z)}$. Specifically, give an input image $x$ and a set of camera poses${\{\pi_1,\pi_2,\cdots,\pi_{n_v}\}}$, $p_m{(z)}$ can be represented as,
\begin{equation}
\label{equation1}
\begin{aligned}
p_m{(z)} = p_d({d^{1:n_d}_{1:n_v}}|x,\pi_{1:n_v}),
\end{aligned}
\end{equation}
where $p_d$ is the distribution of images from $n_d$ different domains. MVDMs train a model $f$ to generate multi-view images with geometric and visual consistency. The model $f$ includes an encoder $E$ that maps the input image $x$ to a latent feature space $z=E(x)$. Then, a UNet denoiser~\cite{ronneberger2015u} with various attention mechanisms is used for denoising. Meanwhile, a pre-trained CLIP model~\cite{radford2021learning} generates embeddings of input image $x$ and camera poses $\pi_{1:n_v}$ to guide the denoising process. Finally, a decoder $D$ maps the latent feature vector back to the initial space $d^{1:n_d}_{1:n_v}=D(z)$.

The key idea of the proposed attack scheme is to optimize imperceptible adversarial noise $\delta$ which can disrupt the inner feature and consistency
across the generated multi-view images simultaneously. The overall framework is illustrated in Figure~\ref{figure2}. Firstly, we extract the latent feature of the adversarial image and design a latent feature erasure loss $L_{FE}$ to drive it away from the distribution of the clean image. Secondly, we randomly sample timestep in each iteration and establish an attention erasure loss $L_{AE}$ to divert the attention of the region of interest to other regions, thereby disrupting the geometric and visual consistency among the generated multi-view
images. Eventually, we combine $L_{FE}$ and $L_{AE}$ to form the final dual erasure loss, then the perturbation can be updated by the gradient descent algorithm. In the following content, we will illustrate the proposed feature and attention dual erasure attack in detail.

\subsection{Feature and Attention Dual Erasure} 

To disrupt the inner feature of each generated multi-view image, we introduce a feature erasure loss in Eq. (\ref{equation2}), which causes the latent features of the adversarial examples $x'$ to deviate from the latent features of the clean images $x$.
\begin{equation}
\label{equation2}
\begin{aligned}
L_{FE}=-(E(x')-E(x))^2.
\end{aligned}
\end{equation}
The feature erasure loss is similar to previous works~\cite{liang2023adversarial,zhu2024watermark} on single image generation tasks, and the following contents mainly present the design method of the attention erasure loss.

\begin{figure}[t!]
\centering
\includegraphics[width=0.7\linewidth]{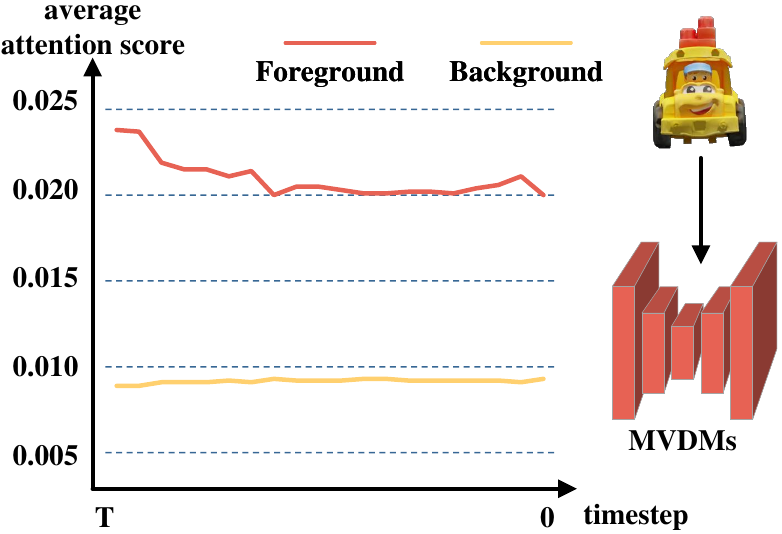}
\caption{
Average attention score received by foreground and background on Wonder3D.
}
\label{figure3}
\end{figure}

As mentioned above, the geometric and visual consistency among the generated multi-view images is attributed to various attention mechanisms in MVDMs, thus we attempt to erasure the attention on the region of interest by reducing the corresponding attention scores.

The representative attention mechanisms employed in MVDMs include self-attention, multi-view attention and cross-domain attention. The attention map can be calculated as,
\begin{equation}
\label{equation4}
\begin{aligned}
m=Softmax(\frac{QK^T}{\sqrt{d}}).
\end{aligned}
\end{equation}

Typically, regions that have higher attention scores are more important to enhance the consistency among multi-view images. Therefore, we conduct a brief analysis of the degree of attention paid to the regions of foreground and background. Considering that the attention maps of lower dimensions contain more semantic information~\cite{hertz2022prompt, chefer2023attend}, we aggregate and average all attention maps of the lowest dimension. As shown in Eq. (\ref{equation3}), the average attention scores of foreground $\Bar{T_f}$ and background $\Bar{T_b}$ can be obtained.
\begin{equation}
\label{equation3}
\begin{aligned}
\Bar{T_f}&=\frac{\sum m_{ij}M'_{ij}}{\sum M'_{ij}}\\
\Bar{T_b}&=\frac{\sum m_{ij}(1-M'_{ij})}{\sum (1-M'_{ij})},
\end{aligned}
\end{equation}
where $M'$ is a binary matrix mask indicating the location of the foreground in attention map. Figure~\ref{figure3} shows the average attention scores received by foreground and background on Wonder3D. We can see that the foreground has higher attention and the attention erasure loss for various attention mechanisms is directly defined as,
\begin{equation}
\label{equation5}
\begin{aligned}
L_{AE}=T_f=\sum m_{ij}M'_{ij}.
\end{aligned}
\end{equation}

In Eq. (\ref{equation5}), the dimensions of $m$ and $M'$ are totally different for various attention mechanisms employed in MVDMs. Specifically, the self-attention mechanism aims to build internal correlation among different regions in each single-view and single-domain image. Therefore, its $K$ and $Q$ have the same dimensions as $(B,D,C)$, where $B$ represents batch size, $D$ represents sequence length and $C$ represents hidden size. The dimension of attention map $m$ for self-attention is $(B, D, D)$. The multi-view attention mechanism aims to build external correlation among different regions across multiple view images. Therefore, the dimension of the original $K$ is firstly reshaped to $(B/n_v,D*n_v,C)$, and then repeated $n_v$ times to $(B,D*n_v,C)$ to match the dimension of $Q$. The dimension of attention map $m$ for multi-view attention is $(B, D, D*n_v)$. Similarly, the cross-domain attention mechanism aims to build external correlation among different regions across multiple domain images. Therefore, the dimension of $K$ is transformed to $(B,D*n_d,C)$. The dimension of attention map $m$ for self-attention is $(B, D, D*n_d)$.

To match the dimension of $m$, we calculate $M'$ by firstly extracting the foreground mask $M$ of the input image. Then we resize it to $(\sqrt{D}, \sqrt{D})$ and reshape it to $(1, D)$ to gain the foreground mask of each row and repeat it $D$ times to gain the foreground mask for each view and domain. Finally, according to different attention mechanisms, we further repeat it to $(D,D)$, $(D,D*n_v)$, $(D,D*N_d)$ respectively for self-attention, multi-view attention and cross-domain attention.

In summary, the abovementioned $L_{FE}$ and $L_{AE}$ are eventually combined to form the dual erasure loss as shown in Eq. (\ref{equation7}). $\alpha$ is the weight to control the balance of the two losses. We use PGD~\cite{madry2017towards} to update the perturbations. The training process is illustrated in Supplementary Material.
\begin{equation}
\label{equation7}
\begin{aligned}
L_{DE}=L_{AE}+\alpha L_{FE}.
\end{aligned}
\end{equation}

It is worth noting that the calculation of attention erasure loss is related to the sampling of time steps. Due to memory constraints, it is challenging to directly optimize the attention loss at all time steps simultaneously. Therefore, we use the Monte Carlo algorithm to randomly sample time steps for optimization.

\begin{figure*}[t!]
\centering
\includegraphics[width=\linewidth]{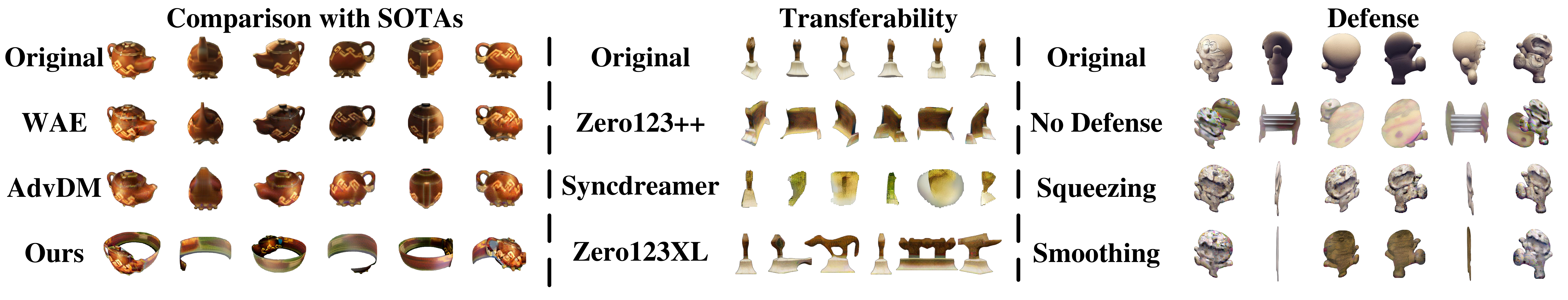}
\caption{
Some visualization results. (\textit{Left})Attack performance of various methods. (\textit{Middle})Transferability on other models. (\textit{Right})Robustness of our method.
}
\label{figure4}
\end{figure*}

\section{Experiments}
% In this section, we first present the experimental settings. Then we conduct extensive experiments under various configurations to assess the effectiveness, robustness, and transferability of the proposed method.
\subsection{Experiments Settings}
\subsubsection{Datasets}
We evaluate our method on the widely used Google Scanned Objects dataset~\cite{downs2022google}. The evaluation dataset is constructed with the same manner as previous researches~\cite{liu2023syncdreamer, long2024wonder3d}. They selected 30 categories of objects from this dataset, including toys, furniture, animals, etc. For each object in the evaluation set, they rendered a 256$\times$256 front view as the input image and other views' images as the ground truth. In addition, to address more complex application scenarios and evaluate the generalizability of our method, they also added some images with diverse styles from the internet to the evaluation set.
\subsubsection{Evaluation metrics}
We evaluate our method from two perspectives. Firstly, for the multi-view images generated by MVDMs, we use two commonly used image quality metrics, namely SSIM and LPIPS. A lower SSIM and a higher LPIPS indicate a greater difference between the generated image and the ground truth. Apart from the quality of the generated multi-view images, we also evaluate the quality of the geometric reconstruction shapes by using the Chamfer Distance (CD) metric. A higher CD indicates a greater difference between the reconstructed shapes and the ground truth.
\subsubsection{Implementation details}
Zero123++~\cite{shi2023zero123++} and Wonder3D~\cite{long2024wonder3d} are two representative SOTA MVDMs and we conduct adversarial attacks on these models. The former uses self-attention and generates multi-view color images, while the latter uses multi-view attention and cross-domain attention and additionally generates normal maps. The 3D reconstruction based on the multi-view images generated by these two models is implemented by One2345++~\cite{liu2024one} and the method introduced in Wonder3D. We use the official pre-trained weights of these models in the following experiments. As for the hyperparameters, the maximum epoch is set to 100, the $\alpha$ used to balance the loss weight is set to 1, the perturbation range $\delta$ is set to 16 and the denoising steps is set to 20. All experiments are conducted on NVIDIA V100 GPUs.

\begin{table}[t]
  % \scriptsize
    % \scriptsize
    \footnotesize
  \caption{
  Objective comparison of attack performance among different methods on Zero123++ and Wonder3D.}  
  \label{table_1}
  \centering
  \begin{tabular}{c|ccc}
       \hline
Attack&\multicolumn{3}{c}{Zero123++}\\
Method & SSIM$\downarrow$ & LPIPS$\uparrow$ & CD$\uparrow$\\
\hline
original & 0.907 & 0.124 & 0.0306\\
WAE & 0.867 & 0.177 & 0.0530\\
AdvDM & 0.849 & 0.181 & 0.0554\\
Ours &  \textbf{0.655} & \textbf{0.405} & \textbf{0.2155} \\
\hline
 Attack&\multicolumn{3}{c}{Wonder3D}\\
Method & SSIM$\downarrow$ & LPIPS$\uparrow$ & CD$\uparrow$\\
\hline
original & 0.917 & 0.084 & 0.0274\\
WAE& 0.860 & 0.184 & 0.0742\\
AdvDM& 0.854 & 0.204 & 0.0757\\
Ours& \textbf{0.781} & \textbf{0.342}  & \textbf{0.1845}\\
\hline
  \end{tabular}
\end{table}

\subsection{Comparison with SOTAs}
As mentioned above, we are the first to consider intellectual property infringement issues arising from
MVDMs and currently there are no specific attack methods for multi-view image generation tasks. Therefore, we compare our method with two SOTA attack methods AdvDM~\cite{liang2023adversarial} and WAE~\cite{zhu2024watermark} designed for single-image generation tasks. WAE used a feature loss to drive the latent feature of the input image to a specific watermark image, while AdvDM utilized a diffusion loss to disrupt the distribution of the latent feature. To compare the attack performance of different methods, we perform AdvDM, WAE, and the proposed method on Zero123++ and Wonder3D individually.

The results are shown in Table~\ref{table_1}, the values in the `original' line mean the quality of the generated multi-view images and the reconstruction shapes without any attack. As can be observed, WAE and AdvDM only cause small variations in SSIM, LPIPS, and CD scores, while our method achieves significant attack effects on all the metrics. Specifically, our method reduces SSIM to 0.655 and increases LPIPS to 0.405 and CD to 0.2155 in Zero123++. These results demonstrate that our method has better attack performance than previous works in terms of the quality of the generated multi-view images and the geometric reconstruction shapes.

Figure~\ref{figure4}(\textit{Left}) further illustrates the subjective comparison of different methods. The upper part shows the color images generated by Zero123++. For each attack method, we present the generated images of six different views. We can see that previous methods only result in slight changes in the surface texture and the outline still remains complete. In contrast, our method generates totally different features and outlines, because the proposed dual erasure attack can greatly disrupt the geometric and visual consistency among multi-view images. More visualization results including reconstruction results based on generated multi-view images can be find in Supplementary Material.

\begin{table}[h]
  % \scriptsize
    % \scriptsize
    \footnotesize
  \caption{
  The ablation study on feature erasure loss $L_{FE}$ and attention erasure loss $L_{AE}$.}  
  \label{table_2}
  \centering

  \begin{tabular}{cc|ccc}
       \hline
\multicolumn{2}{c|}{Losses}&\multicolumn{3}{c}{Zero123++} \\
$L_{FE}$ & $L_{AE}$& SSIM$\downarrow$ & LPIPS$\uparrow$ & CD$\uparrow$\\
\hline
$\times$ & $\times$ & 0.907 & 0.124 & 0.0306\\
$\checkmark$ & $\times$ & 0.848 & 0.228 & 0.0656\\
$\times$ & $\checkmark$ & 0.672 & 0.376 & 0.1975\\
$\checkmark$ & $\checkmark$ & \textbf{0.655} & \textbf{0.405} & \textbf{0.2155}\\
\hline
\multicolumn{2}{c|}{Losses}&\multicolumn{3}{|c}{Wonder3D} \\
$L_{FE}$ & $L_{AE}$& SSIM$\downarrow$ & LPIPS$\uparrow$ & CD$\uparrow$\\
\hline
$\times$ & $\times$ & 0.917 & 0.084 & 0.0274\\
$\checkmark$ & $\times$ & 0.825 & 0.263 & 0.0802\\
$\times$ & $\checkmark$ & 0.787 & 0.319 & 0.1758\\
$\checkmark$ & $\checkmark$ & \textbf{0.781} & \textbf{0.342} & \textbf{0.1845}\\
\hline
  \end{tabular}
\end{table}

\subsection{Ablation Study}
\subsubsection{Impact of different losses}
To demonstrate the impact of our proposed dual erasure attack of features and attention, we firstly conduct an ablation experiment regarding the two losses $L_{FE}$ and $L_{AE}$ presented in Eq. (\ref{equation7}). Specifically, we set the weight of one loss to zero while keeping the other, comparing the results of single erasure (either feature or attention) with those of dual erasure. The results are shown in Table~\ref{table_2}. The second row shows the results of using only $L_{FE}$. In this case, since it only generates some chaotic effects while the consistency of the multi-view images is not disrupted, the image quality does not degrade significantly. The third row shows the results of using only $L_{AE}$. We can see that all the metrics have an obvious variation. The combination of $L_{FE}$ and $L_{AE}$ achieves the best performance as shown in the last row.

% In this case, the image quality can drop significantly, but its effect is still inferior to that of dual erasure. Therefore, it can be seen that our proposed attention erasure loss, while disrupting the consistency of multi-view images, can further degrade image features when combined with feature loss, achieving the desired attack effect.

\begin{table}[h]
  % \scriptsize
    % \scriptsize
    \footnotesize
  \caption{
  The ablation study on different maximum perturbation bounds.}  
  \label{table_3}
  \centering

  \begin{tabular}{c|ccc}
       \hline
\multirow{2}*{Bound}&\multicolumn{3}{c}{Zero123++} \\
& SSIM$\downarrow$ & LPIPS$\uparrow$ & CD$\uparrow$\\
\hline
4 & 0.699 & 0.316 & 0.1593\\
8 & 0.686 & 0.339 &  0.1777\\
16 & 0.655 & 0.405 & 0.2155 \\
32 & 0.623 & 0.418 & 0.2314 \\
\hline
\multirow{2}*{Bound}&\multicolumn{3}{|c}{Wonder3D}\\
& SSIM$\downarrow$ & LPIPS$\uparrow$ & CD$\uparrow$\\
\hline
4 & 0.811 & 0.291 & 0.1635\\
8 & 0.797 & 0.315 & 0.1724\\
16 & 0.781 & 0.342 & 0.1845\\
32 & 0.739 & 0.447 & 0.2059\\
\hline
  \end{tabular}
\end{table}

% \begin{figure*}[t!]
% \centering
% \includegraphics[width=\linewidth]{pic/figure6-transferability.pdf}
% \caption{
% Transferability on other models. The first column shows the generated images on Wonder3D and Zero123++. The other columns show the generated images on other models. For each set of images, the first row represents the images generated from clean images, while the second row represents the images generated from adversarial examples.
% }
% \label{figure6}
% \end{figure*}

\subsubsection{Maximum perturbation bound}
In the field of adversarial attacks, there is often a trade-off between the invisibility of the noise and the effectiveness of the attack. To explore the impact of different perturbation strengths on the attack effectiveness, we conduct experiments with different boundaries to test the attack effectiveness of our method under various boundary constraints. Table~\ref{table_3} shows the results under four different perturbation bounds: 4, 8, 16, and 32. As can be seen, the attack effectiveness gradually improves with the increase in perturbation strength. Since the adversarial noise is visible to some extent when setting the maximum perturbation bound to 32, we set it as 16 in our experiments. It also can be noted that the proposed dual erasure attack achieves better performance even at the setting of the lowest perturbation bound 4 than previous methods when setting the perturbation bound as 16 (As shown in Table~\ref{table_1}).

% Additionally, when the strength is limited, the attack effectiveness can still be maintained at an acceptable level. Therefore, in practical applications, users can adjust the perturbation strength to balance the invisibility of the noise and the effectiveness of the attack.

\subsubsection{Number of denoising steps}
For diffusion models, the number of denoising steps is a crucial factor, as it significantly influences the quality of the images. The training process of our method is also related to the number of denoising steps. Therefore, we further conduct experiments with different numbers of denoising steps to investigate its impact on attack effectiveness. The results are shown in Table~\ref{table_4}. It can be observed that as the denoising steps increase, the effectiveness of the attack decreases slightly. This is mainly because the generation process enables more refined and robust results with larger denoising steps. However, the results indicate that our method always has a satisfactory attack performance on different denoising steps.
\begin{table}[h]
  % \scriptsize
    % \scriptsize
    \footnotesize
  \caption{
  The ablation study on denoising steps.}  
  \label{table_4}
  \centering

  \begin{tabular}{c|ccc}
       \hline
\multirow{2}*{Steps}&\multicolumn{3}{c}{Zero123++} \\
& SSIM$\downarrow$ & LPIPS$\uparrow$ & CD$\uparrow$\\
\hline
10 & 0.628 & 0.418 & 0.2216 \\
20 & 0.655 & 0.405 & 0.2155 \\
30 & 0.670 & 0.383 & 0.2026 \\
40 & 0.669 & 0.385 & 0.1952 \\
\hline
\multirow{2}*{Steps}&\multicolumn{3}{|c}{Wonder3D}\\
& SSIM$\downarrow$ & LPIPS$\uparrow$ & CD$\uparrow$\\
\hline
10 & 0.745 & 0.368 & 0.1920\\
20 & 0.781 & 0.342 & 0.1845\\
30 & 0.808 & 0.327 & 0.1764\\
40 & 0.813 & 0.315 & 0.1677\\
\hline
  \end{tabular}
\end{table}

% \begin{figure*}[t!]
% \centering
% \includegraphics[width=\linewidth]{pic/figure5-defense.pdf}
% \caption{
% Robustness of adversarial examples. The first column shows the multi-view images generated by clean images. The second column shows the images generated by adversarial examples. The third and fourth columns show the images generated by adversarial examples which are preprocessed by defense methods.
% }
% \label{figure5}
% \end{figure*}

\subsection{Transferability on Other Models}
The above experiments verify the effectiveness of the adversarial examples generated by Wonder3D and Zero123++ on these two models. In this subsection, we further explore the transferability of our method on other models, including SyncDreamer~\cite{liu2023syncdreamer} and Zero123XL~\cite{liu2023zero}. Specifically, we directly use the adversarial examples generated by Wonder3D and Zero123++ to attack other models that are unseen during the training process. The subjective results are shown in Figure~\ref{figure4}(\textit{Middle}). As can be seen, the generated multi-view images on SyncDreamer and Zero123XL have been attacked successfully. The textures and outlines are completely destroyed, and these results indicate that our method is applicable to various MVDMs and has good transferability in protecting 3D assets in practical applications.

\subsection{Robustness of Adversarial Examples}
To evaluate the robustness of our method, we adopt two widely used adversarial defense methods on the generated adversarial examples, including squeezing and smoothing~\cite{xu2017feature}. These defense methods are added as a preprocessing procedure before the input. The results are shown in Table~\ref{table_5}, it can be observed that all the metrics only present small variations. Attack effectiveness decreases slightly when facing defense methods. Figure~\ref{figure4}(\textit{Right}) illustrates some subjective results in the case of no defense and two defense methods. We can find that the generated multi-view images still suffer from severe disruption in terms of textures and outlines. This indicates that our method is robust against defense methods.

\begin{table}[h]
  % \scriptsize
    % \scriptsize
    \footnotesize
  \caption{
  The attack performance against defense methods.}  
  \label{table_5}
  \centering

  \begin{tabular}{c|ccc}
       \hline
\multirow{2}*{Defenses}&\multicolumn{3}{c}{Zero123++} \\
& SSIM$\downarrow$ & LPIPS$\uparrow$ & CD$\uparrow$\\
\hline
No defense & 0.655 & 0.405 & 0.2155 \\
Squeezing & 0.697 & 0.344 &  0.1807\\
Smoothing & 0.692 & 0.373 &  0.1896\\
\hline
\multirow{2}*{Defenses}&\multicolumn{3}{|c}{Wonder3D}\\
& SSIM$\downarrow$ & LPIPS$\uparrow$ & CD$\uparrow$\\
\hline
No defense & 0.781 & 0.342 & 0.1845\\
Squeezing & 0.810 & 0.307 & 0.1503\\
Smoothing & 0.798 & 0.322 & 0.1643\\
\hline
  \end{tabular}
\end{table}

\section{Conclusion}
In this work, we propose a novel method to prevent intellectual property infringement caused by MVDMs. Speciafically, a latent feature and attention dual erasure attack method is proposed to disrupt the distribution of latent feature and the consistency across the generated multi-view images simultaneously. Extensive experiments show that our method achieves SOTA attack effectiveness, transferability, and robustness on various MVDMs and defense methods. Therefore, this work provides an effective method to prevent the theft of 3D assets.
\bibliographystyle{IEEEbib}
\bibliography{submission}

\end{document}